%% file: main.tex
\documentclass[runningheads]{llncs}
\usepackage[T1]{fontenc}
\input{preamble}
\usepackage{graphicx,verbatim}
\begin{document}
%
\title{Distilling Photon-Counting CT into Routine Chest CT through Clinically Validated Degradation Modeling}
\titlerunning{\method}
%

\author{Junqi Liu\inst{1} \and
Xinze Zhou\inst{1} \and
Wenxuan Li\inst{1} \and
Scott Ye\inst{2} \and
Arkadiusz Sitek\inst{3} \and
Xiaofeng Yang\inst{4} \and
Yucheng Tang\inst{5} \and
Daguang Xu\inst{5} \and
Kai Ding\inst{6} \and
Kang Wang\inst{2} \and
Yang Yang\inst{2} \and 
Alan L. Yuille\inst{1} \and
Zongwei Zhou\inst{1,6}\thanks{Correspondence to: Zongwei Zhou (\href{mailto:zzhou82@jh.edu}{\texttt{zzhou82@jh.edu}})}
}
\authorrunning{J. Liu et al.}
\institute{
Johns Hopkins University \and
University of California, San Francisco \and
Harvard Medical School \and
Emory University \and
Nvidia \and
Johns Hopkins Medicine
}

  
\maketitle              
\begin{abstract}

Photon-counting CT (PCCT) provides superior image quality with higher spatial resolution and lower noise compared to conventional energy-integrating CT (EICT), but its limited clinical availability restricts large-scale research and clinical deployment. To bridge this gap, we propose \method, a simulated \emph{degradation-to-enhancement} method that learns to reverse realistic acquisition artifacts in low-quality EICT by leveraging high-quality PCCT as reference. Our central insight is to explicitly model realistic acquisition degradations, transforming PCCT into clinically plausible lower-quality counterparts and learning to invert this process. The simulated degradations were validated for clinical realism by board-certified radiologists, enabling faithful supervision without requiring paired acquisitions at scale. As outcomes of this technical contribution, we: (1) train a latent diffusion model on \numofpcct\ PCCTs, using an autoencoder first pre-trained on both these PCCTs and \numofeict\ EICTs from \numofhospital\ hospitals to extract general CT latent features that we release for reuse in other generative medical imaging tasks; (2) construct a large-scale dataset of over \numofct\ publicly available EICTs enhanced to PCCT-like quality, with radiologist-validated voxel-wise annotations of airway trees, arteries, veins, lungs, and lobes; and (3) demonstrate substantial improvements: across external data, \method\ outperforms state-of-the-art image translation methods by 15\% in SSIM and 20\% in PSNR, improves radiologist-rated clinical utility in reader studies, and enhances downstream top-ranking lesion detection performance, increasing sensitivity by up to 15\% and F1 score by up to 10\%. Our results suggest that emerging imaging advances can be systematically distilled into routine EICT using limited high-quality scans as reference. All datasets, code, and models are available at \url{https://github.com/KumaKuma2002/OpenVAE}.


\end{abstract}

\section{Introduction}\label{sec:introduction}

Computed tomography (CT) is a frontline imaging modality for the screening, diagnosis, and longitudinal monitoring of thoracic diseases. In routine clinical practice, CT systems equipped with energy-integrating detectors (EICT) are often constrained by image noise and limited spacial resolution, which contribute to image degradation affects such as partial volume effects and other  artifacts. These limitations can obscure small anatomical  structures, such as distal airways and fine pulmonary vessels, and subtle pathological findings. 

Photon-counting CT (PCCT) is a recent hardware innovation with improved characteristics relative to EICT \cite{shah2025photon}. In contrast to energy integrating detectors, PCCT systems register individual x-ray photons and estimate their energy, providing intrinsic spectral sensitivity while improving dose efficiency and spatial resolution  \cite{van2025photon,varga2025photon}. These technical advances improve the visibility of small anatomical structures and reduce artifacts, with potential benefits for airway analysis, vascular assessment, and lesion characterization in chest CT \cite{chamberlin2023ultra,toth2024neurovascular}. Despite this promise, PCCT remains difficult to implement at scale. PCCT scanners are substantially more expensive than EICT scanners and are concentrated in a limited number of well-resourced academic and tertiaty-care centers, restricting broad clinical access and disproportionately limiting availability in underserved communities \cite{van2025photon,varga2025photon}. 

This limited penetration also impedes large-scale research efforts: high-quality PCCT data are not readily available in the volumes and across the diverse practice settings needed for robust population studies and external validation and development of generalizable models \cite{van2025photon,varga2025photon}. Consequently, most clinical imaging workflows, and most public datasets, remain dominated by EICT \cite{hamamci2024developing}. 
The resulting heterogeneity introduces a second barrier: systematic domain shifts between PCCT and EICT (and across EICT platforms/protocols) complicate direct comparison and model transfer, reducing reproducibility and limiting the generalizability of findings in large-scale multi-institutional research \cite{van2025photon,varga2025photon}. Collectively, this landscape creates a fundamental translational gap: while the clinical and technical benefits of PCCT are increasingly recognized, there remains no scalable strategy to extend comparable image quality and analytic advantages to the substantially larger population imaged with conventional EICT systems. 


A direct solution would be to train an image enhancement model that improves quality of EICT to approximate that of PCCT. However, such a strategy would require paired acquisitions, EICT and PCCT scans obtained in the same patient at scale which is impractical and impossible to achieve due to added cost, scanner time, and operational complexity. More importantly, routinely acquiring two scans introduces avoidable radiation exposure and raises ethical and regulatory concerns. In addition, purely data-driven image translation risk producing unrealistic textures or misleading structures, or subtle anatomical distortions. Even minor such artifacts are unacceptable in clinical imaging and can undermine trust \cite{eulig2024benchmarking,gao2023corediff,liu2025diffusion,tavakoli2025generative,liu2025see,lin2025pixel}. We therefore ask a different question: \emph{can we learn a clinically plausible forward degradation process from PCCT and then train an enhancement method to invert that process in a controlled and auditable way?} 


To address these challenges, we introduce \method, a simulated degradation-to-enhancement method that distills PCCT benefits into EICT using set of \numofpcct\ PCCT scans as reference and more than \numofeict\ EICT scans are examples of lower quality, real-world clinical data. The central concept is to model realistic acquisition degradations that map PCCT to degraded PCCT, thereby emulating the physical and statistical properties observed in EICT. The realism and clinical fidelity of these degradations are prospectively validated by board-certified radiologists to ensure that they reflect authentic acquisition limitations rather than artificial artifacts. An enhancement method is then trained that maps degraded PCCT back to high quality PCCT. This enhancement method can then be applied to any EICT image to provide image enhancement.  Once trained, the enhancement method can be applied to EICT scans to enhance image quality. 

\begin{figure}[htbp]
    \centering
    \includegraphics[width=\linewidth]{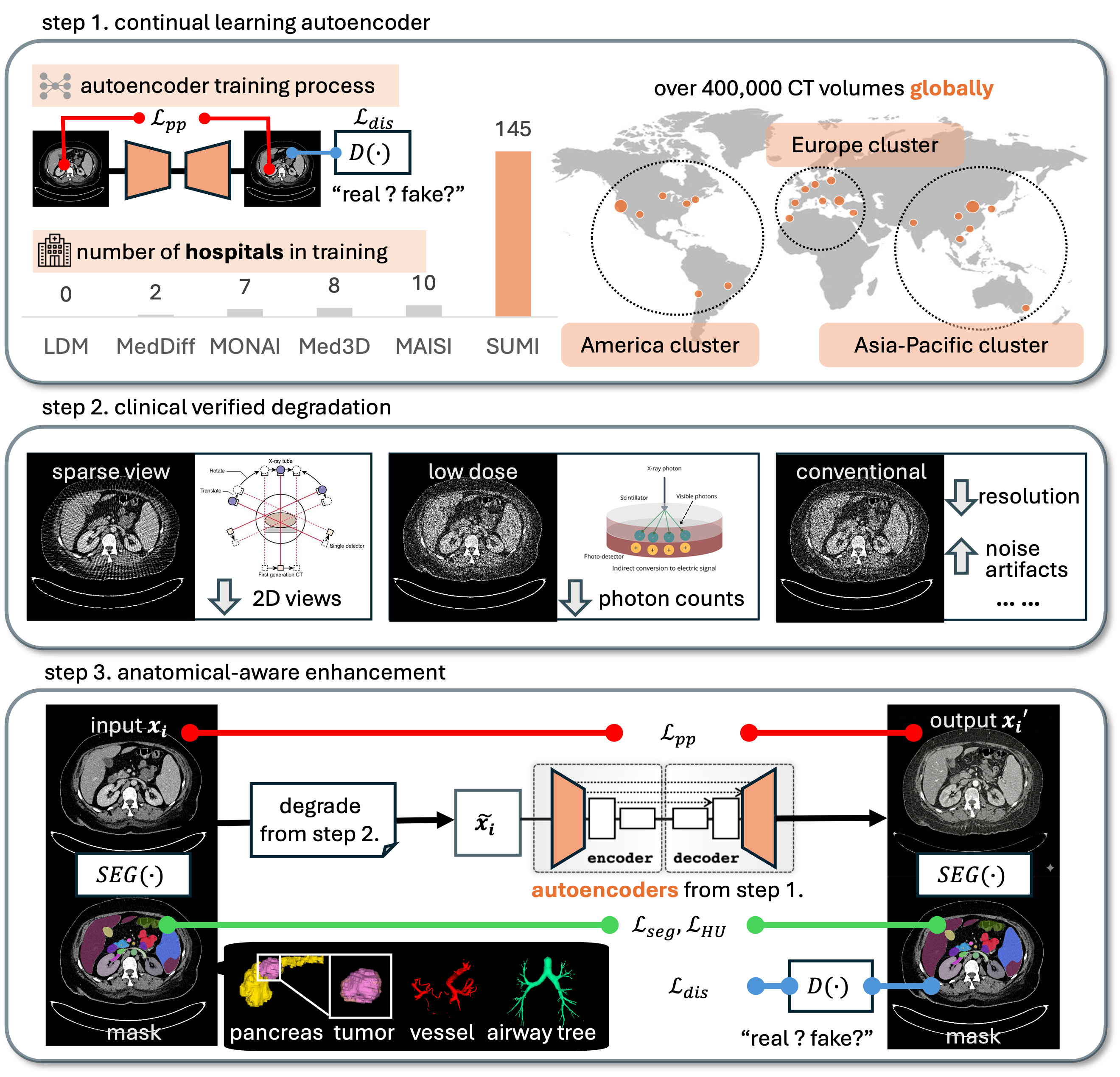}
    \caption{\textbf{Overview of \method.}
    \textbf{(1.)}~A continual learning autoencoder is pre-trained on over 400,000 CT volumes from \numofhospital\ hospitals across \numofcountry\ countries under pixel-wise loss $\mathcal{L}_{pp}$ and adversarial loss $\mathcal{L}_{dis}$, making it the largest and most geographically diverse open-source medical CT autoencoder to date, overpassing prior methods
    (LDM~\cite{rombach2022high}: 0,
    MedDiff~\cite{wang20253d}: 2,
    MONAI~\cite{cardoso2022monai}: 7,
    Med3D~\cite{chen2019med3d}: 8,
    MAISI~\cite{zhao2026maisi}: 10).
    \textbf{(2.)}~A clinical verified degradation simulator transforms high-quality PCCT into three realistic low-quality counterparts: sparse-view (reducing 2D projections), low-dose (reducing photon counts), and conventional (reducing resolution while increasing noise and artifacts), covering the primary sources of EICT degradation.
    \textbf{(3.)}~\method\ takes a degraded input $\tilde{x}_i$ from step~(2.) and passes it through the autoencoder from step~(1.) to produce an enhanced output $x_i'$. Training uses the original PCCT $x_i$ as ground truth under four losses:
    pixel-wise loss $\mathcal{L}_{pp}$ for structural fidelity,
    segmentation loss $\mathcal{L}_{seg}$ and HU consistency loss $\mathcal{L}_{HU}$ to preserve organ boundaries and tissue densities, such as pancreas, tumor, vessel, and airway tree using pre-computed masks $SEG(\cdot)$,
    and adversarial loss $\mathcal{L}_{dis}$ via discriminator $D(\cdot)$ for image realism.
    Once trained, \method\ can enhance any EICT scan to PCCT-like quality without retraining.}
    \label{fig:method}
\end{figure}
\begin{figure}[h]
    \centering
    \includegraphics[width=\linewidth]{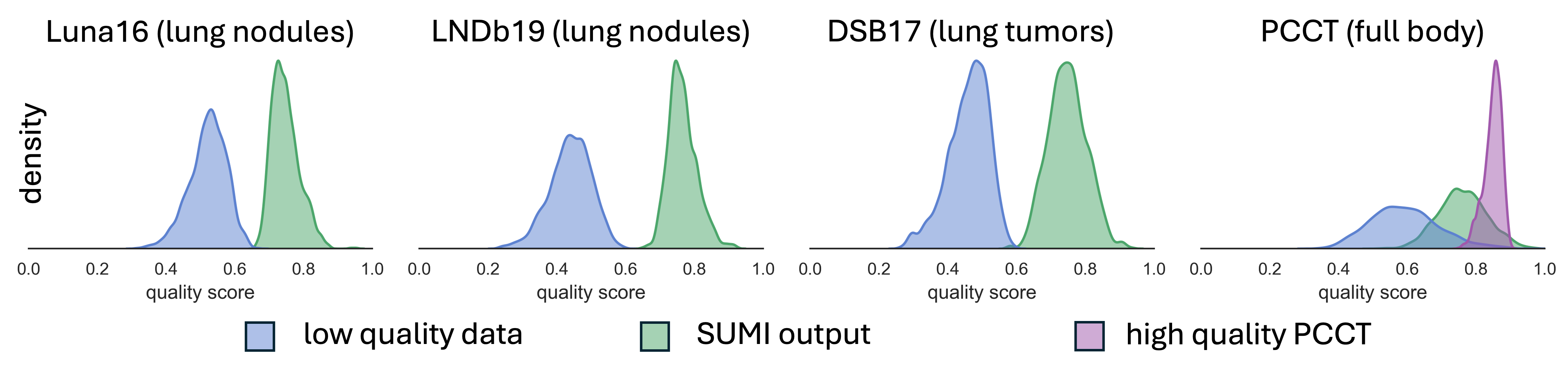}
    \caption{\textbf{\method\ improves chest CT quality across datasets.} 
    Quality score distributions (by a pretrained scorer, 0–1, higher is better) on Luna16~\cite{setio2017validation}, LNDb19~\cite{pedrosa2019lndb}, DSB17~\cite{kuan2017deep}, and a private PCCT dataset. On public datasets, enhanced scans consistently shift toward higher scores, indicating robust quality gains. On PCCT, a controlled degradation-to-enhancement benchmark shows that \method\ narrows the gap to real high-quality PCCT standards.}
    \label{fig:public_data_improve}
\end{figure}

In summary, our contributions are threefold. \textbf{First}, we pre-train an autoencoder on \numofpcct\ PCCT scans and \numofeict\ EICT scans (from \numofhospital\ hospitals; \numofcountry\ countries, see Figure~\ref{fig:method}) to learn general CT representations. Building on this foundation, we train a PCCT-focused diffusion model that generates PCCT-like quality from degraded inputs. \textbf{Second}, we introduce a PCCT-to-EICT degradation simulator that captures realistic acquisition artifacts and validate its clinical realism through review of radiologists, enabling controlled supervision for enhancement (see Figure~\ref{fig:method}). \textbf{Third}, we curate a \numofct\ enhanced CT dataset with PCCT-like image quality and radiologist-validated voxel-wise annotations of airway trees, pulmonary arteries, pulmonary veins, lungs, and lobes. We demonstrate improved image quality (see Table~\ref{tab:baseline}, Figure~\ref{fig:public_data_improve}), higher radiologist-rated clinical utility (see Figure~\ref{fig: corrlation}), and superior downstream lesion detection performance across three independent external datasets (see Table~\ref{tab:reformatted_detection_wide}, Figure~\ref{fig:visualization}). 

Overall, this work outlines a pragmatic strategy for distilling imaging gains achieved with specialized hardware into routine CT workflows using high quality reference scans. It has the potential to facilitate large-scale and multi-institutional research and to generate meaningful clinical impact across diverse care settings. 


\section{Method}\label{sec:method}

Our method is designed for scalability and reuse. We train a latent diffusion model in a learned latent space~\cite{rombach2022high,guo2025text2ct,mao2025medsegfactory,yang2025medical}, and we first pre-train an autoencoder on both the \numofpcct\ PCCT scans and \numofeict\ EICT scans collected from \numofhospital\ hospitals (\S\ref{sec:method_autoencoder}). This autoencoder learns general CT latent features that stabilize diffusion training and can serve as a reusable feature backbone for other generative medical imaging tasks (\S\ref{sec:method_enhancement}).

\subsection{Continual Autoencoder}\label{sec:method_autoencoder}
We propose a large-scale continual-learning autoencoder to address domain shift and catastrophic forgetting in medical imaging. Conventional autoencoders are limited by dataset scale and diversity, leading to poor cross-hospital generalization
~\cite{zhang2020generalizing,perkonigg2021dynamic}.

Our model adopts two strategies: (1) \textbf{Large-scale pretraining} on \numofeict\ CT scans to learn strong anatomical priors for downstream tasks; and (2) a \textbf{Memory Loss module} to preserve previously learned anatomy during sequential adaptation. This enables robust cross-site generalization while allowing hospitals to locally adapt the model without sharing patient data.

\subsection{Degradation-to-Enhancement Method}\label{sec:method_enhancement}
To ensure robust enhancement across heterogeneous CT data, we design a degradation-to-enhancement method that explicitly models real-world acquisition artifacts and learns to reverse them in a controlled manner. 

\smallskip\noindent\textbf{CT Degradation}
High-quality PCCT scans are degraded to simulate common clinical artifacts and generate EICT-like appearances, as verified by an experienced radiologist. We apply three degradation strategies. 
\textbf{Sparse View} reduces projection numbers in Radon space, inducing streak artifacts:
\[
\mathbf{p}_{\text{sparse}} = \mathcal{S}(\mathbf{p}), 
\quad 
\hat{\mathbf{x}}_{\text{sparse}} = \mathcal{R}(\mathbf{p}_{\text{sparse}}).
\]
\textbf{Low Dose} decreases photon counts and models signal-dependent Poisson noise before reconstruction:
\[
\hat{\mathbf{p}} \sim \text{Poisson}(\alpha \mathbf{p}), 
\quad 
\hat{\mathbf{x}}_{\text{low}} = \mathcal{R}(\hat{\mathbf{p}}).
\]
\textbf{Conventional Degradation} applies spatial downsampling with Gaussian and Poisson noise injection to mimic reduced resolution and electronic noise in standard CT scanners.

\smallskip\noindent\textbf{CT Enhancement}
We train a latent diffusion model (LDM) in the autoencoder latent space so that a degraded $\tilde{x}$ maps to an enhanced output $x'$ with the paired PCCT $x$ as ground truth, as in step~(3) of Figure~\ref{fig:method}. After training, the same model enhances routine EICT toward PCCT-like quality without retraining.

Training uses the four losses summarized in Figure~\ref{fig:method}. \textbf{Pixel-wise loss} $\mathcal{L}_{pp} = \| x' - x \|_1$ encourages structural fidelity to PCCT quality. \textbf{Segmentation loss} $\mathcal{L}_{seg} = \text{CE}(\text{SEG}(x'), \text{SEG}(x))$ preserves organ boundaries and tissue densities using pre-computed masks $\text{SEG}(\cdot)$. \textbf{HU consistency loss} $\mathcal{L}_{HU} = \sum_{c} \| \mu_c(x') - \mu_c(x) \|_1$ matches mean Hounsfield units $\mu_c$ within each segmented region $c$. \textbf{Adversarial loss} $\mathcal{L}_{dis} = -\mathbb{E}[\log D(x')]$ with discriminator $D(\cdot)$ improves realism.

\section{Experiment}\label{sec:experiment}

\begin{table}[t]
\centering
\scriptsize
\setlength{\tabcolsep}{3pt}
\caption{\textbf{\method\ improves image quality across degradations.} 
On \numoftestdata\ external patients, our method surpasses the second-best method by up to +35.9\% SSIM and +19.5\% PSNR under sparse-view, low-dose, conventional, and mixed settings. Baselines are ordered by ascending average performance.
}
\label{tab:baseline}
\begin{tabular}{lcccccccc}
\toprule
\multirow{2}{*}{method} & \multicolumn{2}{c}{sparse view} & \multicolumn{2}{c}{low dose} & \multicolumn{2}{c}{conventional} & \multicolumn{2}{c}{mixed} \\
\cmidrule(lr){2-3} \cmidrule(lr){4-5} \cmidrule(lr){6-7} \cmidrule(lr){8-9}
 & SSIM & PSNR & SSIM & PSNR & SSIM & PSNR & SSIM & PSNR \\
\midrule
SR3~\cite{saharia2022image}    & 46.2 & 8.7  & 42.0 & 8.8  & 44.4 & 8.6  & 47.0 & 8.7  \\
Pix2Pix~\cite{isola2017image}  & 43.6 & 11.9 & 60.0 & 22.6 & 60.8 & 18.1 & 62.9 & 22.7 \\
Swin2SR~\cite{conde2022swin2sr}& 62.1 & 26.7 & 59.9 & 28.0 & 29.0 & 24.9 & 66.8 & 26.4 \\
NLM~\cite{buades2005non}       & 73.2 & 28.4 & 72.3 & 30.2 & 40.9 & 26.1 & 76.9 & 27.0 \\
SRGAN~\cite{ledig2017photo}    & 56.8 & 23.2 & 85.5 & 33.0 & 82.8 & 32.4 & 63.1 & 26.5 \\
NEED~\cite{gao2025noise}       & 67.4 & 27.8 & 83.1 & 34.1 & 80.3 & 33.1 & 69.5 & 28.2 \\
\midrule
\textbf{\method} & \textbf{91.6} & \textbf{33.0} & \textbf{92.8} & \textbf{37.3} & \textbf{88.2} & \textbf{35.0} & \textbf{90.2} & \textbf{33.7} \\
\rowcolor[gray]{0.95} \textbf{$\Delta$} 
& \textbf{\textcolor{DeepGreen}{+35.9\%}} 
& \textbf{\textcolor{DeepGreen}{+18.7\%} }
& \textbf{\textcolor{DeepGreen}{+8.5\%} }
& \textbf{\textcolor{DeepGreen}{+9.4\%} }
& \textbf{\textcolor{DeepGreen}{+6.5\%}} 
& \textbf{\textcolor{DeepGreen}{+5.7\%}} 
& \textbf{\textcolor{DeepGreen}{+17.3\%}} 
& \textbf{\textcolor{DeepGreen}{+19.5\%}} \\
\bottomrule
\end{tabular}
\end{table}

\noindent\textbf{Datasets and evaluation.} 
Our autoencoder is pre-trained on \numofeict\ EICT scans from \numofhospital\ hospitals, while \method\ is trained on \numofpcct\ high-quality PCCT scans. Image quality (SSIM/PSNR) is evaluated on \numoftestdata\ held-out private PCCT cases, and downstream detection is assessed on Luna16~\cite{setio2017validation}, LNDb19~\cite{pedrosa2019lndb}, and DSB17~\cite{kuan2017deep}. We further release an enhanced open-source set of \numofct\ public EICT scans enhanced to PCCT-like quality (Table~\ref{tab:public_datasets}) with radiologist-validated voxel-wise annotations of airways, arteries, veins, lungs, and lobes.

\begin{figure}[ht]
    \centering
    \includegraphics[width=\linewidth]{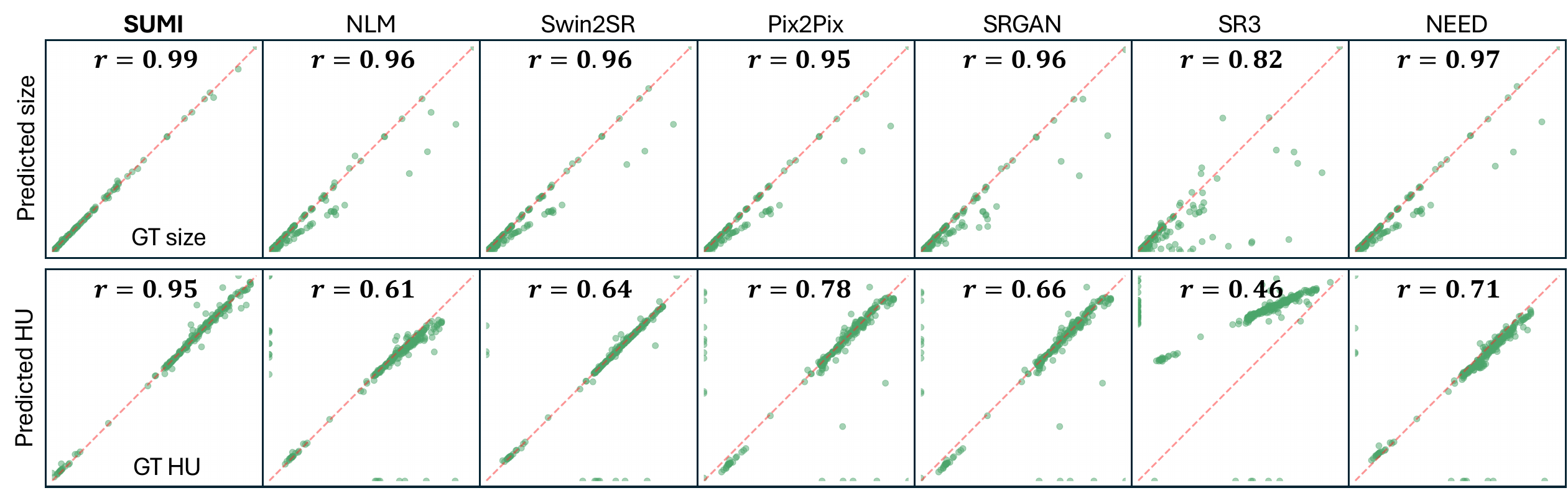}
    \caption{\textbf{\method\ preserves anatomical structure and tissue density.} 
    Pearson correlation ($r$) between ground truth and enhanced CT measurements across all organs shows that \method\ maintains anatomical accuracy and HU consistency. Organ masks are obtained with VISTA3D~\cite{he2025vista3d}.}
    \label{fig: corrlation}
\end{figure}

\begin{table*}[ht]
    \centering
    \scriptsize
    \setlength{\tabcolsep}{8pt}
    \caption{\textbf{Our method processes and enhances a massive, multi-source cohort of \numofct\ public chest CT scans.} Cohorts are listed in alphabetical order.}
    \label{tab:public_datasets}
    \begin{tabular}{l cccc}
        \toprule
        \textbf{dataset} & DSB17~\cite{kuan2017deep} & LIDC-IDRI~\cite{armato2011lung} & LNDb19~\cite{pedrosa2019lndb} & LTRC~\cite{bartholmai2006lung} \\
        \textbf{scans ($N$)} & 1,596 & 1,018 & 324 & 1,496 \\
        \midrule
        \textbf{dataset} & Luna16~\cite{setio2017validation} & MIDRC~\cite{whitney2023longitudinal} & NLST~\cite{national2011national} & RSNA-STR~\cite{colak2021rsna} \\
        \textbf{scans ($N$)} & 854 & 10,496 & 422 & 1,110 \\
        \bottomrule
    \end{tabular}
\end{table*}

\noindent\textbf{Baseline and implementation.} 
We select representative methods from four major categories: traditional filtering (NLM~\cite{buades2005non}), Vision Transformers (Swin2SR~\cite{conde2022swin2sr}), GAN-based models (Pix2Pix~\cite{isola2017image}, SRGAN~\cite{ledig2017photo}), and diffusion models (SR3~\cite{saharia2022image}, NEED~\cite{gao2025noise}). These models are primarily developed for natural images; adapting them to medical CT requires substantial pipeline redesign and full retraining. Moreover, many advanced medical enhancement methods remain closed-source, limiting reproducible comparison.

\begin{figure}[ht]
    \centering
    \includegraphics[width=\linewidth]{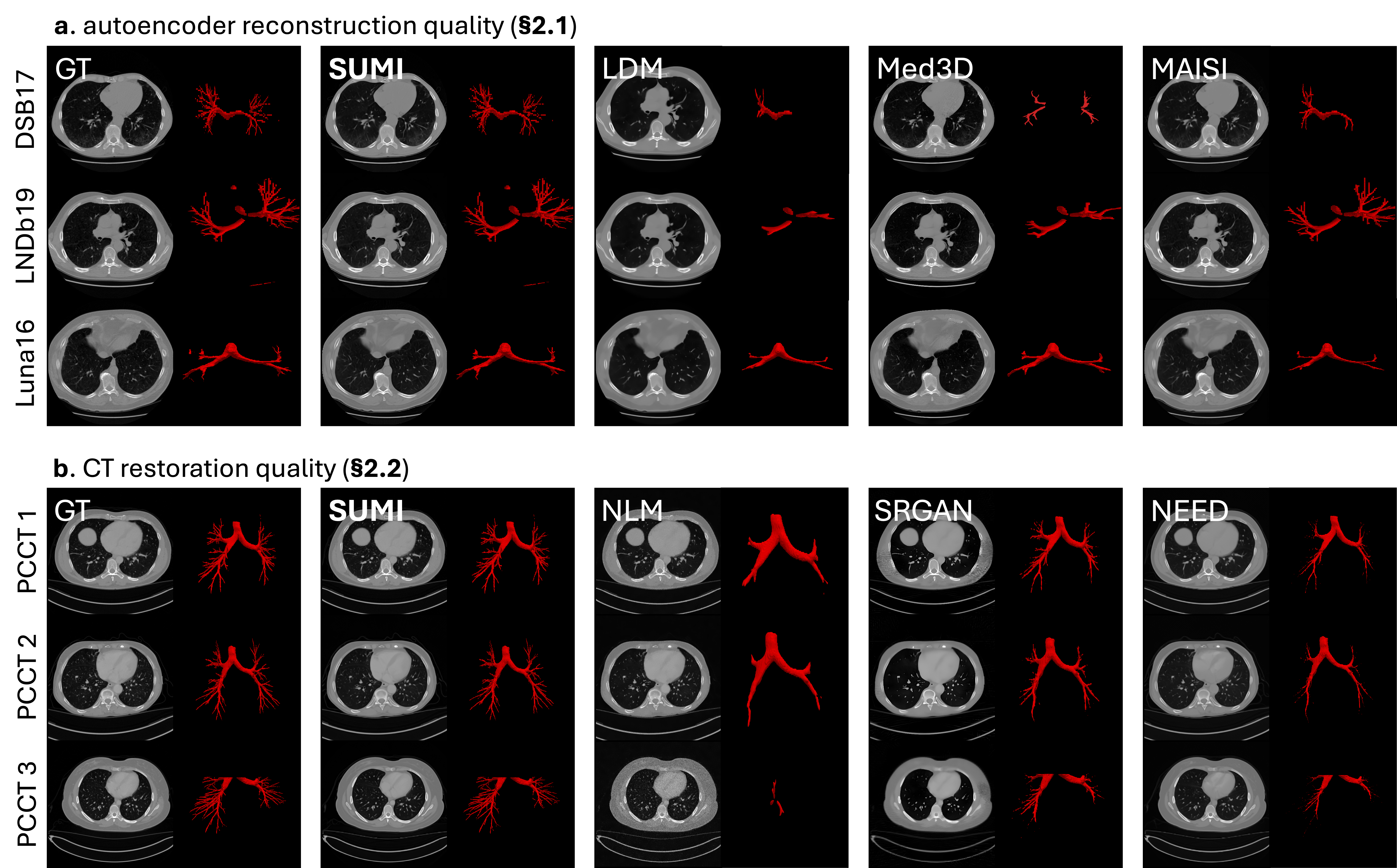}
    \caption{
    \textbf{\method\ demonstrates superior generalization and enhancement quality.}
    For each example, the left image shows the CT slice and the right shows the airway tree segmentation.
    (a) Cross-dataset evaluation: compared with autoencoders trained with limited medical data scale (LDM, Med3D, MAISI), \method\ better preserves fine airway topology under domain shift.
    (b) PCCT enhancement with ground truth: compared with strong baselines (NLM~\cite{buades2005non}, SRGAN~\cite{ledig2017photo}, NEED~\cite{gao2025noise}), \method\ maintains structural integrity and small airway branches more faithfully.
    }
    \label{fig:visualization}
\end{figure}

\smallskip\noindent\textbf{Image Quality and Enhancement} We compare our method with diverse baselines (Table~\ref{tab:baseline}). Across sparse-view, low-dose, conventional, and mixed degradations, \method\ consistently outperforms all competitors, exceeding the second-best model by up to +35.9\% SSIM and +19.5\% PSNR. As shown in Figure~\ref{fig:visualization}, our method effectively suppresses noise and streak artifacts, producing sharp CT slices that closely resemble the ground truth.

\begin{table*}[h]
    \centering
    \scriptsize
    \renewcommand{\arraystretch}{1.1} 
    \setlength{\tabcolsep}{2pt}
    \begin{threeparttable}
    \caption{\textbf{\method\ improves downstream detection across three independent chest CT cohorts.} 
    When integrated into open-source SOTA baselines, our enhancement consistently increases performance on Luna16~\cite{setio2017validation}, LNDb19~\cite{pedrosa2019lndb}, and DSB17~\cite{kuan2017deep}, achieving up to +15.2\% F1 and +10.5\% AUC. Baselines are trained on standard-quality CT, while ``+ \method'' denotes the same architectures retrained using our PCCT-like enhanced data.}
    \label{tab:reformatted_detection_wide}
    
    \begin{tabular}{l cccc c cccc c cccc}
        \toprule
        & \multicolumn{4}{c}{\textbf{Luna16~\cite{setio2017validation} (N=222)}} && \multicolumn{4}{c}{\textbf{LNDb19~\cite{pedrosa2019lndb} (N=224)}} && \multicolumn{4}{c}{\textbf{DSB17~\cite{kuan2017deep} (N=198)}} \\
        & \multicolumn{4}{c}{lung nodule, \textit{MONAI}\textsuperscript{1}~\cite{cardoso2022monai}} 
        && \multicolumn{4}{c}{lung nodule, \textit{MONAI}\textsuperscript{1}~\cite{cardoso2022monai}} 
        && \multicolumn{4}{c}{lung tumor, \textit{grt123}\textsuperscript{2}~\cite{liao2019evaluate}} \\
        \cmidrule{2-5} \cmidrule{7-10} \cmidrule{12-15}
        method & Sen. & Spec. & F1 & AUC && Sen. & Spec. & F1 & AUC && Sen. & Spec. & F1 & AUC \\
        \midrule
        
        baseline 
        & 75.8 & 88.4 & 81.6 & 88.0 && 
        61.5 & 86.9 & 55.2 & 74.0 && 
        78.4 & 82.1 & 80.2 & 87.0 \\
        
        \quad + \method 
        & 84.5 & 88.4 & 87.2 & 92.5 && 
        75.0 & 89.2 & 70.4 & 84.5 &&
        88.2 & 86.4 & 87.3 & 91.8 \\
        
        \rowcolor[gray]{0.95} \quad \textbf{$\Delta$} 
        & \textcolor{DeepGreen}{\textbf{+8.7}} & +0.0 & \textcolor{DeepGreen}{\textbf{+5.6}} & \textcolor{DeepGreen}{\textbf{+4.5}} && 
        \textcolor{DeepGreen}{\textbf{+13.5}} & \textcolor{DeepGreen}{\textbf{+2.3}} & \textcolor{DeepGreen}{\textbf{+15.2}} & \textcolor{DeepGreen}{\textbf{+10.5}} && 
        \textcolor{DeepGreen}{\textbf{+9.8}} & \textcolor{DeepGreen}{\textbf{+4.3}} & \textcolor{DeepGreen}{\textbf{+7.1}} & \textcolor{DeepGreen}{\textbf{+4.8}} \\
        
        \bottomrule
    \end{tabular}
    \begin{tablenotes}
        \item \textsuperscript{1}MONAI: First-place open-source SOTA model for \href{https://project-monai.github.io}{LUNA16/LNDb19 challenges}.
        \item \textsuperscript{2}grt123: First-place SOTA model for  \href{https://github.com/lfz/DSB2017}{DSB2017 challenge}.
    \end{tablenotes}
    \end{threeparttable}
\end{table*}

\smallskip\noindent\textbf{Anatomical Fidelity and Tissue Consistency}
Medical generative models must preserve physical fidelity. Figure~\ref{fig: corrlation} reports Pearson correlation ($r$) of organ size and Hounsfield Units (HU) between ground truth and enhanced CTs, segmented with VISTA3D~\cite{he2024vista3d}. The results show that our method preserves anatomical boundaries and tissue densities without structural hallucination.

\smallskip\noindent\textbf{Downstream Clinical Performance.}
Standardizing image quality directly improves diagnostic performance. As shown in Table~\ref{tab:reformatted_detection_wide}, replacing standard-quality CT with our enhanced scans increases F1 by up to +15.2\% and AUC by +7.5\%. Quality score distributions further demonstrate a consistent shift toward real PCCT standard

\smallskip\noindent\textbf{Ablation Study.}
\begin{table}[b]
\centering
\scriptsize
\caption{\textbf{Ablation of degradation simulations in \method.} 
Leave-one-out experiments under identical settings show that removing sparse-view or low-dose simulation degrades performance on an external test set (\numoftestdata\ scans), confirming their role in robust generalization.}
\label{tab:ablation}
\begin{tabular}{lcccccccc}
\toprule
\multirow{2}{*}{method} & \multicolumn{2}{c}{sparse view} & \multicolumn{2}{c}{low dose} & \multicolumn{2}{c}{conventional} & \multicolumn{2}{c}{mixed} \\
\cmidrule(lr){2-3} \cmidrule(lr){4-5} \cmidrule(lr){6-7} \cmidrule(lr){8-9}
 & SSIM & PSNR & SSIM & PSNR & SSIM & PSNR & SSIM & PSNR \\
\midrule

w/o. sparse view
& 71.9 & 30.2
& 90.0 & 35.2
& 87.2 & 35.3 
& 73.2 & 30.1 
\\
w/o. low dose
& 72.7 & 30.2 
& 82.6 & 28.0 
& 84.3 & 32.9 
& 76.2 & 31.5 
\\
w/o. conventional
& 75.8 & 27.0 
& 86.0 & 33.0 
& 85.3 & 32.0 
& 81.9 & 32.4 \\
\midrule
\textbf{all degrades} & \textbf{91.6} & \textbf{33.0} & \textbf{92.8} & \textbf{37.3} & \textbf{88.2} & \textbf{35.0} & \textbf{90.2} & \textbf{33.7}\\
\bottomrule
\end{tabular}
\end{table}
A leave-one-out ablation study (Table~\ref{tab:ablation}) validates our method. Omitting any single degradation strategy during training causes a distinct performance drop in that specific scenario, confirming all simulations are essential for robust clinical generalization.

\section{Discussion and Conclusion}\label{sec:discussion_conclusion}

\noindent\textbf{Limitation.} Our study has several limitations: (1) Evaluation is limited to chest CT; extending to other anatomies is planned. Note that chest CT quality degradation is more pronounced than abdominal/pelvic CT, making this focus more impactful. (2) Current implementation uses 2D slices rather than full 3D volumes. We validated z-dimension continuity by processing consecutive slices, but full 3D implementation is deferred due to computational constraints and left for future work.

\smallskip\noindent\textbf{Conclusion.} We present a novel AI-driven method that learns from real PCCT scans to enhance routine low-quality CT scans to PCCT-like image quality. Our method simulates realistic degradation processes to train an enhancement method that improves low-quality CT images to high-quality PCCT-like standards. We demonstrate substantial improvements in image quality and clinical utility across multiple datasets, validated by reader studies with board-certified radiologists. This work demonstrates a transformative paradigm: emerging imaging advances achieved through expensive, specialized hardware (PCCT scanners) can now be democratized and effectively distilled into routine clinical CT scanners using limited high-quality reference scans and artificial intelligence. This eliminates the need for costly hardware upgrades and enables hospitals worldwide to achieve PCCT-quality imaging from their existing CT infrastructure. Our datasets, code, and models will be publicly available.

\begin{credits}
\smallskip\noindent\textbf{\ackname} This work was supported by the Lustgarten Foundation for Pancreatic Cancer Research and the National Institutes of Health (NIH) under Award Number R01EB037669. We would like to thank the Johns Hopkins Research IT team in \href{https://researchit.jhu.edu/}{IT@JH} for their support and infrastructure resources where some of these analyses were conducted; especially \href{https://researchit.jhu.edu/research-hpc/}{DISCOVERY HPC}. We thank Jaimie Patterson for writing a \href{}{news article} about this project. Paper content is covered by patents pending.

\subsubsection{\discintname}
The authors declare no competing interests.
\end{credits}
%
%
%
\clearpage
\bibliographystyle{splncs04}
\bibliography{refs,zzhou}

\end{document}

%% file: preamble.tex
\usepackage{float}
\usepackage{pifont}
\usepackage{footnote}
\usepackage{enumitem}
\usepackage{bm}
\usepackage{arydshln}
\usepackage{booktabs}
\usepackage{multicol}
\usepackage{multirow}
\usepackage{color}
\usepackage{xcolor}     
\usepackage{colortbl}
\usepackage{soul}
\usepackage{bbding}
\usepackage{makecell}
\usepackage{mathtools}
\usepackage{imakeidx}
\usepackage{amssymb}
\usepackage{graphicx}
\usepackage{amsmath}
\usepackage{threeparttable}
\definecolor{citecolor}{HTML}{0071BC}
\definecolor{linkcolor}{HTML}{ED1C24}
\usepackage[colorlinks,
            anchorcolor=red,
            citecolor=citecolor, 
            linkcolor=linkcolor,
            ]{hyperref}
\makeindex
\usepackage{arydshln}
\usepackage{lipsum}
\usepackage[toc]{multitoc}
\usepackage[edges]{forest}
\usepackage[normalem]{ulem}

\usepackage{bbding}
\usepackage[most]{tcolorbox}

\usepackage{algorithm}
\usepackage{algorithmic}

\usepackage{minitoc}
\usepackage[toc,page,header]{appendix}

\definecolor{orchid}{rgb}{0.85, 0.44, 0.84}
\definecolor{rubinred}{rgb}{0.82, 0.0, 0.28}
\definecolor{flagship}{rgb}{0, 255, 0}
\definecolor{radiologist}{rgb}{0.50, 0.50, 1}

\definecolor{YT}{HTML}{002FA7}

\newcommand{\method}{{\fontfamily{ppl}\selectfont
SUMI}}

\newcommand{\numofct}{17,316}
\newcommand{\numofeict}{405,379}
\newcommand{\numofpcct}{1,046}

\newcommand{\numofhospital}{145}
\newcommand{\numofcountry}{19}
\newcommand{\numoftestdata}{446}

\definecolor{citecolor}{HTML}{0071BC}
\definecolor{linkcolor}{HTML}{ED1C24}
\definecolor{iblue}{rgb}{0.06, 0.75, 1.0}
\definecolor{iorange}{RGB}{255,170,120}
\definecolor{LightBlue}{HTML}{ADD8E6}
\definecolor{orchid}{rgb}{0.85, 0.44, 0.84}
\definecolor{rubinred}{rgb}{0.82, 0.0, 0.28}
\definecolor{flagship}{rgb}{0.93, 0.06, 0.41}
\definecolor{radiologist}{rgb}{0.50, 0.50, 1}

\definecolor{DeepGreen}{RGB}{48,173,59}
\definecolor{LRed}{RGB}{230,88,34}

\newcolumntype{P}[1]{>{\centering\arraybackslash}p{#1}}
\newlength\savewidth
  